# Impact of Level 2/3 Automated Driving Technology on Road Work Zone Safety

Zhepu Xu[1*], Ziyi Song[1], Yupu Dong[1], Peiyan Chen[2]

[1]Faculty of Railway Transportation, Shanghai Institute of Technology, Fengxian, Shanghai, China, 201418

[2]College of Computer Science and Technology, Zhejiang University, Xihu, Hangzhou, China, 310007

*Corresponding author: xuzhepu@126.com

## Abstract

As China's road network enters the maintenance era, work zones will become a common sight on the roads. With the development of automated driving, vehicles equipped with Level 2/3 automated driving capabilities will also become a common presence on the roads. When these vehicles pass through work zones, automated driving may disengage, which can have complex effects on traffic safety. This paper explores the impact of Level 2/3 automated driving technology on road safety in high-speed highway work zone environments. Through microscopic traffic simulation method and using full-type traffic conflict technique, factors such as market penetration rate (MPR), traffic volume level, disengagement threshold, and driver takeover style are studied to understand their impact on work zone safety. The study found that the impact of automated driving technology on work zone safety is complex. Disengagement of automated vehicles in work zones reduces the proportion of vehicles that can maintain automated driving status. If takeover is not timely or adequate, it can easily lead to new traffic conflicts. Different factors have varying degrees of impact on work zone safety. Increasing MPR helps reduce the occurrence of single-vehicle conflicts, but it also increases the possibility of multi-vehicle conflicts. Therefore, future research and improvement directions should focus on optimizing the disengagement detection and takeover mechanisms of automated driving systems.

Key Words: Level 2/3 automated driving; Disengagement; Takeover; Work zone safety; Simulation; Safety assessment

# 1 Introduction

With the road network in China entering the maintenance era, maintenance and repair operations will become more frequent, and work zones will become a norm on highways. However, work zones can interfere with the normal operation of traffic flow and easily trigger traffic accidents, posing serious safety hazards. According to the data on work zone accidents in FARS (Fatality Analysis Reporting System) database from 2008 to 2022 by the U.S. National Highway Traffic Safety Administration[1], there were totally 99,830 fatal crashes, 1,976,300 injury-only crashes and 5,100,124 property-damage-only crashes. This shows the severity of work zone safety situation.

In the past decade, we have witnessed the rapid development of automated driving. Vehicles with certain automated driving capabilities have appeared on the road network, and the market penetration rate (MPR) is increasing. They show different characteristics from traditional vehicles, bringing opportunities and challenges to the work zone safety. Work zone accidents are caused by human factors in 82.7% of cases, so automated driving technology is expected to significantly improve the work zone safety level. In particular, when the MPR of fully automated vehicles in the work zone is 100%, it is expected to achieve "zero accidents"[2]. However, in reality, Level 5 automated driving is facing huge challenges in technology, safety, economy and other aspects, and the commercialization time scale is unknown. According to the data of the Ministry of Industry and Information Technology (MIIT)[3], as of the first half of 2023, the new car sales volume of passenger vehicles with Level 2 automated driving function in China has reached 42.4%. In November 2023, four departments including MIIT jointly issued a notice on the "Launch of the Pilot Program for the Approval and Running on the Road of Intelligent Driving Vehicles", formally allowing Level 3 and 4 vehicles to access roads. At present, BYD, AVATR, Deepal, Mercedes-Benz, ARCFOX, BMW, IM and other automobile companies have been awarded Level 3 road test licenses. The next few years will mark a key period for accelerated application of Level 3 automated driving. By contrast, Level 4 vehicles are currently only operating in some test zones, and if they are to be widely used on public roads, more time is needed. It can be predicted that in a long period to come, Level 2/3 vehicles will become the main form of automated vehicles. Their operation on the expressway will gradually become a norm, and their encounter with the work zones will also become a common scenario. Unlike Level 5 fully automated vehicles, Level 2/3 vehicles' automated driving may disengage when passing through work zones, requiring human intervention. This kind of disengagement reduces the proportion of vehicles that can maintain automated driving in traffic flow. If manual takeover is not sufficient and minimum risk maneuver (MRM) mode is triggered, it may easily cause traffic conflicts and thus become a new risk factor.

Therefore, the impact of automated driving technology on road work zone traffic safety is unknown.

This study takes Level 2/3 automated driving vehicles as the research object and explores its impact on traffic safety in the work zone of the expressway. Through this study, it aims to provide theoretical basis for proposing new traffic control methods in work zones, reduce more serious work zone safety issues caused by automated driving, and reduce life and property losses. Although this study focuses on the work zone of the expressway, the relevant results can also provide a reference for other road grades.

The rest of the arrangements are as follows: The second part for literature review, the third part introduces the research idea and methods, fourth part to experimental design and give the results of experiments, fifth part summarized and proposed a future study.

# 2 Literature review

## 2.1 Safety assessment method for work zones

The existence of work zones may interfere with the normal traffic flow, and there are serious hidden dangers. Safety assessment is very important to ensure the safety of the work zones[4]. In recent years, safety assessment methods based on traffic conflict technology (TCT) have attracted more and more attention due to their significant advantages in timeliness[5, 6].

The influential definition of traffic conflict is based on the spatial-temporal proximity: A traffic conflict is an observable situation in which two or more road users approach each other in space and time to such an extent that there is a risk of collision if their movements remain unchanged[5]. According to this definition, scholars have put forward many indicators for quantifying the proximity of road users, which are used to characterize the severity of traffic conflicts. These indicators are mainly classified into three categories[6]: indicators based on temporal proximity (such as TTC, Time to collision), indicators based on spatial proximity (such as PSD, Proportion of stopping distance) and derived indicators deduced from the above two types of indicators (such as DRAC, Deceleration rate to avoid the crash).

However, this definition only focuses on the detection of two-vehicle or multi-vehicle conflicts, thus it is not suitable for assessing single-vehicle conflicts[6-8]. According to the data[1], within the work zones, in addition to two-vehicle and multi-vehicle accidents, single-vehicle accidents are also serious and important accident types with high fatality rates. In addition, the departments often choose to carry out construction during relatively low traffic periods and are promoting night construction to avoid traffic volume as much as possible, which makes it difficult to observe two-vehicle or multi-vehicle traffic conflicts. Meanwhile, single-vehicle accidents at night are even more serious[9]. Therefore, in this scenario, the problem that existing TCTs cannot detect single-vehicle conflicts is particularly prominent.

In previous research[7], the authors proposed a method to detect single-vehicle conflicts by referring to another classic definition of traffic conflict based on evasive behavior. According to this definition, a traffic conflict is a phenomenon in which a road user must take evasive behavior (e.g., lane changing, braking, etc.) to avoid collision. By automatically segmenting and identifying microscopic vehicle behaviors and determining evasive actions, evasive behaviors can be detected. Furthermore, by integrating the SSAM (Surrogate Safety Assessment Model) method, a method to detect all types of traffic conflicts was proposed. Based on this method for detecting all types of traffic conflicts, the authors further improved their assessment method[10] and proposed a comprehensive assessment index that considers both the likelihood and severity of conflicts, called the equivalent number of traffic conflicts. The authors adopted the Level of Safety Service (LOSS) theory to construct an assessment standard suitable for work zones. This enables a fast safety assessment for work zones.

## 2.2 Automated driving and work zone safety

The emergence of automated driving technology has also triggered researchers' thinking about the safety of work zones. According to a study on work zone accidents, 82.7% of accidents are caused by human factors, and automated driving technology is expected to significantly reduce the occurrence of such traffic accidents. Given the complexity of automated driving technology, most literature has focused on fully automated vehicles. By adjusting the MPR of fully automated vehicles within traffic flow, researchers can investigate the impact of automated driving technology on traffic safety[2, 11-15]. For instance, Abdulsattar et al. constructed an agent-based traffic simulation framework to study the effects of varying traffic volumes and MPRs on surrogate safety indicators such as TTC. Their research found that as the MPR of automated vehicles increases, the safety level of work zones also improves. However, to achieve significant results, the required MPR is associated with traffic volume. The greater the traffic volume, the higher the MPR required. When the MPR reaches 100%, it is anticipated that a target of "zero accidents" could be achieved[2].

The discussion surrounding automated driving and work zone safety holds profound implications. As the MPR of automated vehicles gradually increases, it is expected that the safety conditions in work zones will improve significantly, thereby effectively addressing road safety bottlenecks. However, the above researches have notable limitations, primarily reflected in the

following two aspects:

(1) Existing studies primarily focus on fully automated driving vehicles. However, as mentioned in the introduction, Levels 2/3 will dominate the road for an extended period. Unlike fully automated driving vehicles, L2 and L3 vehicles may experience disengagements while passing through work zones, leading to new issues. First, disengagement reduces the proportion of vehicles capable of maintaining automated driving capabilities. Second, if the takeover is not timely or sufficient, this will further disrupt traffic flow and exacerbate traffic conflicts. Consequently, existing researches cannot clearly determine whether L2/L3 automated driving can enhance work zone safety, necessitating further investigation.

(2) In assessing work zone safety, most existing studies apply TCT using classic safety assessment indicators, e.g., TTC, DRAC, etc.. While these indicators can rapidly assess work zone safety, they have certain limitations. They are able to assess conflicts involving two or more vehicles but neglect the occurrence of single-vehicle conflicts. However, single-vehicle conflicts occur frequently within work zones and require greater attention.

## 2.3 Disengagement and takeover of automated driving

(1) Disengagement of automated driving

Disengagement of automated driving refers to the phenomenon where the driver or safety operator must manually intervene to control the vehicle during the operation of the automated driving system, due to technical malfunctions detected by the system or for safety considerations, leading to the exit from automated driving mode[16]. According to the national standard "Classification of Automobiles' Driving Automation" (GB/T 40429-2021), both Level 2 and Level 3 automated driving systems are susceptible to disengagement, requiring timely human takeover.

Data from accidents involving automated driving vehicles on public roads indicate that many incidents are related to disengagement, particularly those resulting in driver or pedestrian casualties. Research shows that, on average, one traffic accident occurs for every 178 disengagement events[17]. The issue of disengagement has gained attention from relevant authorities; the California Department of Motor Vehicles (DMV) mandates that all manufacturers testing automated vehicles in their pilot program must report disengagement data. This data includes information on the time, location, and detailed descriptions of the disengagement events. Since 2014, over 160,000 disengagement records have been accumulated, along with more than 3 million miles of automated driving mileage and over 100 accident reports. Researchers have conducted studies based on this data set, focusing on factors leading to disengagement[18-21]and the establishment of disengagement causation models[16, 22, 23].

(2) Takeover of automated driving

The national standard "Classification of Automobiles' Driving Automation" (GB/T 40429-2021) stipulates that for Level 2 and Level 3 vehicles, human drivers must take over control when disengagement occurs. However, the actual takeover process is not always timely or seamless, potentially leading to different outcomes (also referred to as takeover performance) in varied scenarios, which ultimately impacts traffic flow. Researchers have explored the factors influencing driver takeover outcomes, evaluation metrics for takeover results, and related models[24].

- Factors influencing driver takeover outcomes

Factors influencing driver takeover outcomes can be categorized into three levels: driver-related, traffic environment-related, and automated driving system-related[24]. Specifically, driver-related factors include non-driving-related tasks, fatigue, and individual characteristics. Traffic environment-related factors encompass weather conditions and complex traffic flows. For the automated driving system, key influencing factors include the mode and timing of takeover requests.

- Evaluation metrics for takeover outcomes

Evaluation metrics for takeover outcomes aim to quantify the effectiveness of takeover from multiple dimensions, such as safety, comfort, and smoothness. Common evaluation indicators include the time of hand contact with the steering wheel, the time of foot contact with the pedals, steering response time, and takeover time[24]. To overcome the limitations of relying solely on a single metric for evaluation, various comprehensive assessment methods have been proposed, such as Takeover Controllability rating[25], Takeover Performance Score[26], and Takeover Performance Index[27].

- Models for takeover outcomes

Takeover outcome models quantitatively analyze influencing factors and predict the driver's takeover results. Researchers have discussed several models for takeover outcomes, which can be broadly categorized into three types: classical statistical models, machine learning models, and structural equation models[24].

# 3 Methodology

This study employs microscopic traffic simulation and statistical analysis methods to explore the impact of Level 2/3 automated driving on traffic safety in work zones. L2 and L3 automated driving differ significantly from manual driving; in addition to showing distinct following and lane-changing behaviors during automated phases[28-30], a critical feature is the occurrence of disengagements requiring manual takeover, which can influence the behavior of automated vehicles. Therefore, a key issue in this research is to accurately represent the behaviors of L2 and L3 automated vehicles within microscopic traffic simulation software.Moreover, the influence of automated vehicles on the safety of work zones is affected by numerous factors, making it essential to consider a comprehensive range of variables as another critical aspect of this study. Lastly, as noted in the literature review, the traffic flow characteristics in work zones differ from those on regular road segments; thus, how to comprehensively assess the safety of traffic flow will be the third critical issue addressed in this study. The research follows the methodological framework depicted in Figure 1.

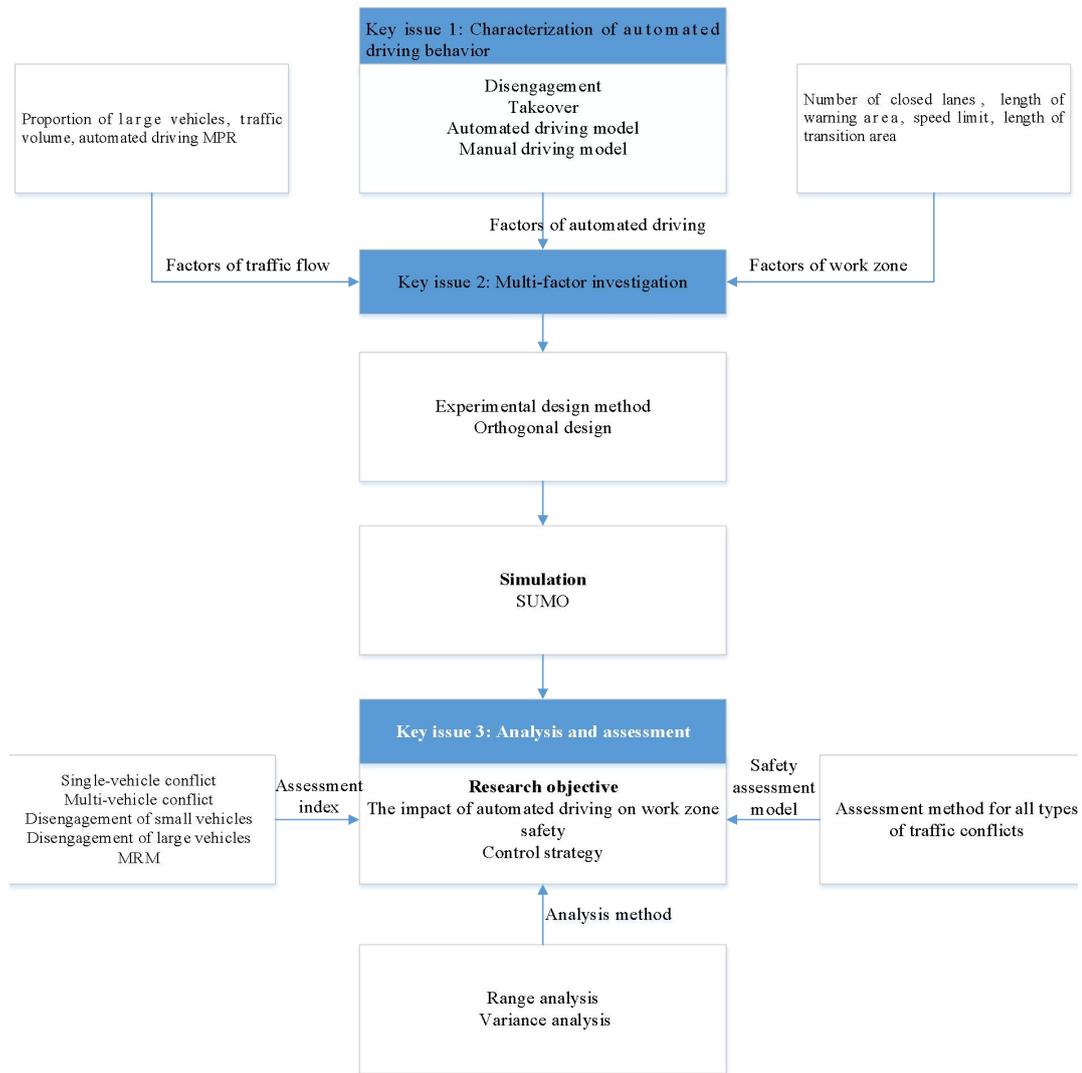

Figure 1 Methodological framework of this study

## 3.1 Representation of L2/L3 automated driving behavior

Currently, simulation is a vital method in automated driving research, and accurately representing L2 and L3 automated driving behavior is a key focus of this study. L2 and L3 automated driving encompasses automated driving mode, manual driving mode, and the transition of control (ToC) between the two. The simulation software employed in this research is SUMO (Simulation of Urban Mobility) [31], thus the selection of the aforementioned models will take into account the latest advancements in SUMO related research.

Considerable research has been conducted on both automated and manual driving modes[28-30]. In this study, model selection is as follows: when the vehicle operates in automated driving mode, the car-following behavior is modeled using the built-in Adaptive Cruise Control (ACC) model of SUMO[32-34], while the lane-changing behavior utilizes the built-in LC2013 model[35]. When the vehicle is in manual driving mode, the car-following behavior is represented using the Krauss model provided by SUMO[36], and the lane-changing behavior uses the LC2013 model.

The transition process is a challenging and critical aspect of representing L2 and L3 automated driving behavior. Significant contributions have been made by researchers such as Lücken and Mintsis[37, 38], whose proposed models provide a solid foundation for this study.

As illustrated in Figure 2, the takeover process for Level 3 vehicles is depicted. When disengagement occurs, the system issues a Take-over Request (TOR) to alert the driver to prepare for takeover. If the driver responds quickly, they can successfully regain control of the vehicle. However, the driver requires a brief period to re-establish situational awareness, during which there may be a temporary decline in their driving performance. Following a short adjustment period, the driver's driving abilities will return to normal. Conversely, if the driver fails to complete the takeover within the designated time frame, the vehicle will initiate a MRM, gradually bringing the vehicle to a full stop at a constant deceleration rate.

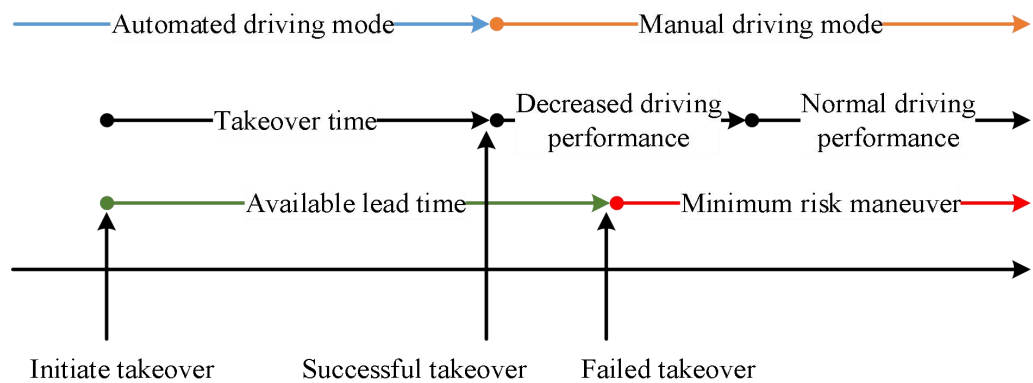

Figure2 Schematic diagram of the takeover process for L3 automated driving

The model for the decline in manual driving performance following a takeover, along with the subsequent recovery, can be expressed using Equation 1. Let $t_0$ represent the moment of takeover, with the driver's performance denoted as $A_0$ (where $A_0$ ranges from 0 to 1). After the takeover, the driver's performance gradually recovers at a fixed rate r, resulting in performance at time t being expressed as $A_0+rt$. During the takeover process, if the driver maintains focused attention, performance may not decline. Therefore, the driver's performance at any given moment can be represented as the minimum value between 1 and $A_0+rt$.

$$A(t_0+t) = Min(1.0, A_0+rt) \tag{1}$$

The model specifies a reserved time for the driver to take over control. If the driver fails to take over in a timely manner, the MRM is triggered, causing the vehicle to decelerate at a constant rate until it comes to a complete stop. Thus, whether the vehicle enters the MRM mode depends on the reserved time and the driver's reaction time. Generally, the reserved time is fixed (for example, 10 seconds), and the deceleration of the vehicle in the MRM is also constant (for example, 3.0 m/s²). The driver's reaction time follows a truncated normal distribution[10, 39]. Consequently, the probability of a vehicle entering the MRM and the duration of the MRM can also be calculated. According to the research findings of Lücken, L., Mintsis, E., et al.[37], the probability of MRM occurrence ranges from 7.7% to 16.2%, while the duration of MRM can vary from 0 to 5 seconds.

Additionally, to facilitate a smoother takeover process, a Gap Control Mode has been introduced, allowing the vehicle to continuously adjust in order to achieve an optimal headway. On one hand, this mechanism enables the vehicle to maintain a coherent and smooth headway under the current control mode; on the other hand, it ensures that the vehicle can also maintain a coherent and smooth headway in the target control mode. This mechanism is particularly beneficial during the transition from a small headway mode to a large headway mode, as it helps prevent the vehicle from executing emergency braking maneuvers. More importantly, this mechanism allows the vehicle to maintain a greater headway during ToC, making it easier for the driver to take over control.

The final question to address is when a TOR occurs. Research conducted by Lücken, L., Mintsis, E., et al. provides an online interface that can trigger a TOR through TraCI. Additionally, a dynamic TOR triggering mechanism is available, which automatically determines whether a TOR is necessary based on the vehicle's surrounding environment. For instance, when a vehicle needs to merge onto a congested road, this scenario may be too complex for an automated driving system. The automatic triggering of a TOR can be configured using the dynamicToCThreshold parameter, which represents the time threshold (in seconds) for the vehicle to continue driving without changing lanes. A TOR is triggered when the remaining distance to the obstacle ahead is less than $Dist_r$. The calculation for $Dist_r$ is shown in Equation (2), where currentSpeed is the vehicle's current running speed, and MRMDist is the distance the vehicle would travel while braking from the current speed to a complete stop.

$$Dist_r = dynamicToCThreshold \times currentSpeed + MRMDist \qquad (2)$$

Wherein, $MRMDist = 0.5 \times currentSpeed^2 / MRMBrakeRate$

## 3.2 Safety assessment method based on all types of traffic conflict technology

In previous research[7], the authors established an all types of traffic conflict detection and safety assessment method applicable to work zones (as illustrated in Figure 3). This method integrates two definitions of traffic conflict: one based on evasive behavior and the other on proximity. The proximity-based detection method, which draws on the SSAM approach and conducts conflict analysis using trajectory files outputted from simulation software, can identify both two-vehicle and multi-vehicle conflicts. In contrast, the evasive behavior-based detection method employs micro-behavioral analysis to extract and identify evasive actions, enabling the detection of single-vehicle, two-vehicle, and multi-vehicle conflicts. By analyzing the overlapping aspects of these two definitions, it is possible to effectively distinguish single-vehicle conflicts, thereby facilitating the detection of all types of traffic conflicts. To comprehensively assess the likelihood and severity of accidents triggered by traffic conflicts, the authors further proposed an integrated evaluation metric—Unit Equivalent Traffic Conflict Number (UETCN) [10]. In this metric, the probability of an accident occurring is derived through probabilistic reasoning, while severity is determined using vehicle energy collision theory. Additionally, to assess the safety level of work zones, the authors introduced a safety assessment method based on the Level of Safety Service (LOSS)[10]. This entire set of methods provides a comprehensive and rapid safety assessment for work zones. For more detailed information, readers are referred to the referenced papers[7, 10].

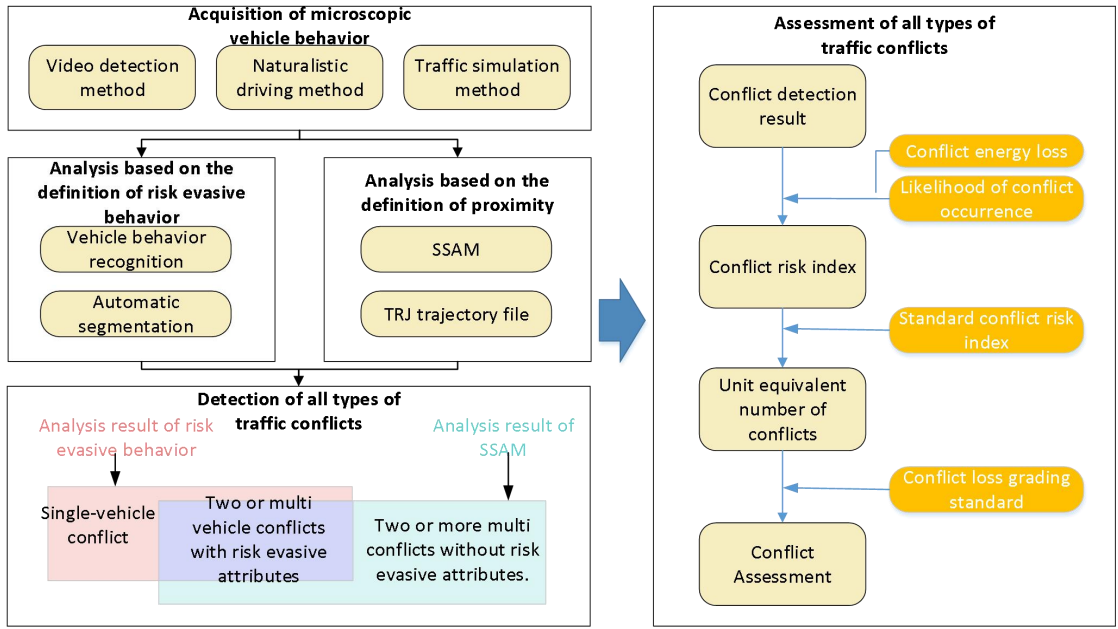

Figure 3 Detection and assessment of all types of traffic conflicts

## 3.3 Factors selection for study

Traffic safety in work zones is influenced by a variety of factors, including traffic flow variables, work zone management factors, environmental conditions, and more. While it is not feasible to replicate all potential factors in micro-simulation, it is possible to simulate the primary influencing factors. As discussed in Section 3.1, the characterization of automated driving behavior illustrates that numerous factors impact the behavior of automated vehicles, ultimately affecting traffic flow and, consequently, traffic safety in work zones. Furthermore, our previous research on safety in work zones under traditional traffic flow conditions identified several critical factors that influence safety, including traffic volume, the proportion of large vehicles, the length of the warning area, and speed limits in the work zone[10]. Specifically, speed limit of the warning area, traffic volume, and the proportion of large vehicles all demonstrated significant positive correlations with UETCN. Additionally, the length of the warning area was found to have a significant negative impact on the equivalent conflict number of two and multi-vehicles. These relationships can be modeled using a Poisson distribution, with correlation coefficient exceeding 0.65.

Moreover, prior studies have indicated that disengagement, takeover, and MPR of automated vehicles can also affect safety in work zones.

(1)Disengagement

As described in Section 3.1, the disengagement model adopted in this study enables automatic disengagement[37]. This occurs when conditions such as excessively high traffic density or overly complex environments render the automated vehicle unable to cope, prompting it to initiate disengagement. The model is controlled by a key parameter, the dynamicToCThreshold, indicating that the selection of the disengagement threshold value has a significant impact on the occurrence of disengagement events.

(2) Takeover

Disengagement and takeover are significant characteristics of Level 2/Level 3 automated driving vehicles. As illustrated in Figure 2, when disengagement occurs, the driver's takeover behavior can have varying impacts on the vehicle and the overall state of traffic flow. Section 3.1 provides a detailed introduction to the vehicle takeover process, highlighting multiple parameters that influence the outcomes of the takeover. This study references the work of Lücken, L., Mintsis, E., et al[37], categorizing takeover styles into three types: aggressive, normal, and conservative. Each takeover style corresponds to different parameter values. Generally, aggressive drivers maintain a higher level of attentiveness, resulting in minimal performance decline during disengagement. In contrast, conservative drivers exhibit a significant decline in performance during disengagement, while normal drivers fall somewhere in between.

The main control parameters related to takeover style, along with their meanings and ranges of values, are summarized in Table 1.

Table 1 Adjustable parameters of the takeover style model

| Parameter Name | Description | Value Range |
|---|---|---|
| sigma | The driver's imperfection (between 0 and 1). | normal(0.2,0.5) |
| tau (s) | The driver's desired (minimum) time headway | normal(0.6, 0.5) |
| decel (m/s2) | The deceleration capability of vehicles. | normal(3.5, 1.0) |
| accel (m/s2) | The acceleration capability of vehicles. | normal(2.0, 1.0) |
| emergencyDecel (m/s2) | The maximum deceleration capability of vehicles. | 9.0 |
| lcAssertive | Willingness to accept lower front and rear gaps on the target lane. | 1.3 |

*Note：In the table, the cell marked as "normal" indicates that the parameter follows a normal distribution, with the mean and standard deviation of the distribution listed in parentheses.

（3）MPR of automated vehicles

Numerous scholars have investigated the impact of MPR on traffic safety. By adjusting the MPR of fully automated vehicles within traffic flow, researchers can study the effects of automated driving technology on traffic safety[2, 11-15]. For instance, Abdulsattar et al. developed an agent-based traffic simulation framework to examine the influence of varying traffic volume levels and different MPRs on surrogate safety measures such as TTC[2].

Consequently, this study considers several factors, including traffic volume, the proportion of large vehicles, the length of warning area, speed limit in work zone, disengagement, takeover, and the MPR of automated vehicles.

## 3.4 Establishment of Evaluation Metrics

This study builds on previous work and references the findings of Lücken, L., Mintsis, E., et al[37], employing indicators including conflict counts, disengagement counts, and MRM counts for safety assessment. Conflicts are further categorized into single-vehicle conflicts and multi-vehicle conflicts, while disengagements are classified by vehicle type into small vehicle (SV) disengagements and large vehicle (LV) disengagements. Similarly, MRM counts are differentiated according to vehicle type into SV MRMs and LV MRMs. The basic meanings of each indicator and their significance are described as follows:

●Conflict Counts (distinguished into single-vehicle, two and multi-vehicle conflicts): A single-vehicle conflict entails a collision between a vehicle and a fixed object or leaving the roadway within the work zone. The higher the number of single-vehicle conflicts, the less safe the situation is perceived to be. Additionally, an increase in two and multi-vehicle conflicts also indicates a deterioration in safety.

● Disengagement Counts (distinguished into SV and LV): This indicator reflects instances where automated driving cannot adequately manage dynamic tasks, resulting in passive disengagements. Generally, a higher number of disengagements signifies greater scene complexity and greater challenges for automated vehicles, or indicates that automated driving technology is less mature.

● MRM Counts (distinguished into SV and LV): When disengagements occur during automated driving, it may be due to scene complexity or takeover style, resulting in the inability to take over promptly and leading the vehicle into a MRM mode. This scenario serves as an indicator of the complexity of the takeover situation to some extent.

# 4 Experimental design and results

## 4.1 Experiment description

This study selects a real work zone as the experimental subject. The work zone is located on the S20 Shanghai Outer Ring Expressway, as shown in Figure 4. The S20 is a bidirectional eight-lane highway with a normal speed limit of 80 km/h. The maintenance task involves repairing potholes in the rightmost first and second lanes; therefore, these lanes will be cordoned off during the construction process, while the third and fourth lanes remain open to traffic.

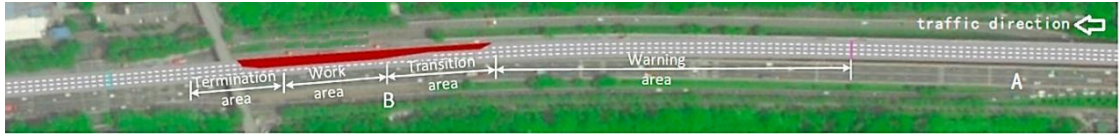

Figure 4 Location and spatial distribution of the case study work zone

This study employs a simulation method using the widely utilized SUMO software within the field of automated driving for the simulation platform. A work zone model is established and calibrated on the SUMO platform, incorporating the ToC model developed by Lücken, L., Mintsis, E., et al. Furthermore, a comprehensive traffic safety assessment method based on the all types of traffic conflict technology is constructed using the data extracted from the SUMO simulation. To assess the severity of traffic conflicts under various scenarios, a complete network model of S20 is developed, along with established evaluation criteria. This approach allows for an analysis of the impact of different factors, including traffic composition, traffic volume levels, and takeover behavior styles, on traffic safety in the work zone.

## 4.2 Experimental design

The study explores the influence of various disengagement thresholds (as detailed in Table 2), takeover behavior styles (as outlined in Table 3), traffic volumes (as indicated in Table 4), the proportion of large vehicles (as shown in Table 5), the lengths of warning area(as presented in Table 6), speed limits in the work zone (as described in Table 7), and traffic compositions (as detailed in Table 8) on the safety levels of the work zone. An orthogonal table is utilized to arrange the experiments, as shown in Table 9. To minimize experimental measurement errors, each experimental group is conducted twice, resulting in a total of 36 trials. Upon completion of the experiments, conflict indicators are extracted based on the all types of traffic conflict technology. Additionally, disengagement counts and MRM counts are obtained from the SUMO output for subsequent result analysis.

Table 2 Disengagement thresholds

| Number | Disengagement thresholds（s） |
|---|---|
| 1 | 5 |
| 2 | 10 |
| 3 | 15 |

Table 3 Parameter settings for takeover styles

| Parameter Name | Aggressive takeover | Normal takeover | Conservative takeover |
|---|---|---|---|
| initialAwareness | normal(0.7,0.3) | normal(0.5,0.3) | normal(0.3,0.3) |
| recoveryRate | normal(0.2,0.1) | normal(0.2,0.1) | normal(0.2.0.1) |

| mrmDecel(m/s2) | 3.0 | 3.0 | 3.0 |
|---|---|---|---|

*Note：   In the table, the cell marked as "normal" indicates that the parameter follows a normal distribution, with the mean and standard deviation of the distribution listed in parentheses. For the meaning of the parameters, please refer to the SUMO official documentation at https://sumo.dlr.de/wiki/ToC_Device.

### Table 4 Traffic volume

| Number | Traffic level | Traffic volume(veh/h) | V/C |
|---|---|---|---|
| 1 | Low traffic volume | 2500 | 0.42 |
| 2 | Medium traffic volume | 3500 | 0.58 |
| 3 | High traffic volume | 4500 | 0.75 |

### Table 5 Proportion of large vehicles

| Number | Proportion of large vehicles |
|---|---|
| 1 | 2% |
| 2 | 22% |
| 3 | 50% |

### Table 6 Length of warning area

| Number | Length of warning area （m） |
|---|---|
| 1 | 600 |
| 2 | 800 |
| 3 | 1000 |

### Table 7 Speed limit in work zone

| Number | Speed limit （km/h） |
|---|---|
| 1 | 40 |
| 2 | 60 |
| 3 | 80 |

### Table 8 Traffic composition

| Number | Proportion of human-driven vehicles （%） | Proportion of automated vehicles （%） | Number | Proportion of human-driven vehicles （%） | Proportion of automated vehicles （%） |
|---|---|---|---|---|---|
| 1 | 100 | 0 | 4 | 40 | 60 |
| 2 | 80 | 20 | 5 | 20 | 80 |
| 3 | 60 | 40 | 6 | 0 | 100 |

### Table 9 Orthogonal design table

| Number | Disengagement thresholds | Parameter settings for takeover styles | Traffic volume | Proportion of large vehicles | Length of warning area | Speed limit in work zone | Traffic composition (AV mpr) |
|--------|--------------------------|----------------------------------------|----------------|------------------------------|------------------------|--------------------------|------------------------------|
| 1 | 5s | aggressive | 2500 | 2% | 600 | 40 | 0% |
| 2 | 5s | aggressive | 2500 | 22% | 800 | 60 | 20% |
| 3 | 5s | aggressive | 2500 | 50% | 1000 | 80 | 40% |
| 4 | 5s | normal | 4500 | 2% | 800 | 80 | 60% |
| 5 | 5s | normal | 4500 | 22% | 1000 | 40 | 80% |
| 6 | 5s | normal | 4500 | 50% | 600 | 60 | 100% |
| 7 | 10s | normal | 3500 | 2% | 600 | 40 | 40% |
| 8 | 10s | normal | 3500 | 22% | 800 | 60 | 0% |
| 9 | 10s | normal | 3500 | 50% | 1000 | 80 | 20% |
| 10 | 10s | conservative | 2500 | 2% | 800 | 80 | 100% |
| 11 | 10s | conservative | 2500 | 22% | 1000 | 40 | 60% |
| 12 | 10s | conservative | 2500 | 50% | 600 | 60 | 80% |
| 13 | 15s | aggressive | 3500 | 2% | 800 | 80 | 80% |
| 14 | 15s | aggressive | 3500 | 22% | 1000 | 40 | 100% |
| 15 | 15s | aggressive | 3500 | 50% | 600 | 60 | 60% |
| 16 | 15s | conservative | 4500 | 2% | 600 | 40 | 20% |
| 17 | 15s | conservative | 4500 | 22% | 800 | 60 | 40% |
| 18 | 15s | conservative | 4500 | 50% | 1000 | 80 | 0% |

## 4.3 Experimental results and analysis

For the orthogonal experiments, data analysis typically employs intuitive analysis method (also known as range analysis) and variance analysis method. Range analysis is used to assess the magnitude of the effect of different factors on the indicators, where a larger range indicates a greater impact, allowing for a preliminary understanding of their influence patterns. Variance analysis is utilized to determine whether specific factors have an effect on the indicators; a sufficiently small P-value suggests a clear impact and serves as a valuable complement to range analysis.

The results of the range analysis are presented in Figure 5. For instance, Figure 5(a) illustrates the variation in the number of single-vehicle conflicts, multi-vehicle conflicts, and total conflicts across different factors and their levels. Each factor is represented by a cluster of curves: the blue line denotes the number of single-vehicle conflicts as the factor levels change, the green line indicates the number of multi-vehicle conflicts, and the red line shows the total number of conflicts. Notably, due to the significant differences in the ranges of each indicator, three vertical axes are utilized for clearer representation. It can be observed that when the disengagement threshold changes from Level 1 to Level 2 and then to Level 3, the numbers of single-vehicle conflicts are 5212, 6820, and 6254, respectively, while the numbers of multi-vehicle conflicts are 32803, 7470, and 25309, and the total conflict counts are 38016, 14290, and 31564.

The results of the variance analysis are shown in Table 10, which presents the P-values for different factors affecting each indicator. According to statistical principles, when the P-value is less than 0.01 (indicated by ** in the table), there is a 99.7% confidence that the factor has a significant impact on the indicator; when the P-value is less than 0.05 (indicated by * in the table), there is a 95% confidence in the significance of the factor's impact. Taking single-vehicle conflicts as an example, the P-values for disengagement threshold, traffic volume, and MPR are all less than 0.01, indicating a significant impact of these factors on single-vehicle conflict indicators. Furthermore, the P-values for takeover style parameters and the proportion of large vehicles are less than 0.05, suggesting that these factors also have a significant influence. Conversely, since the P-values for warning area length and work zone speed limit exceed 0.05, we cannot conclude that these factors have a significant impact on single-vehicle conflicts.

（1）Range Analysis

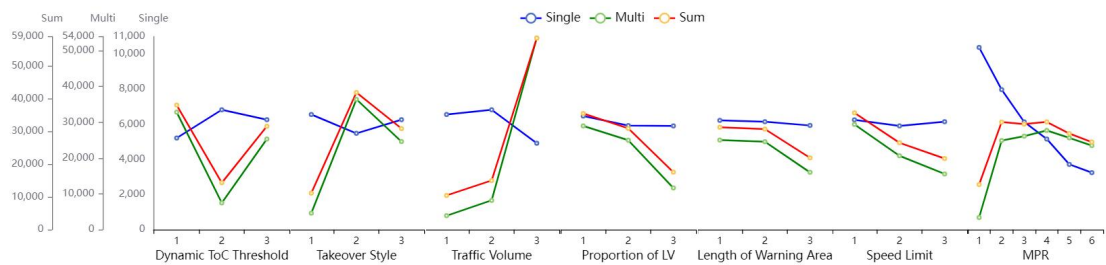

(a) Traffic conflict

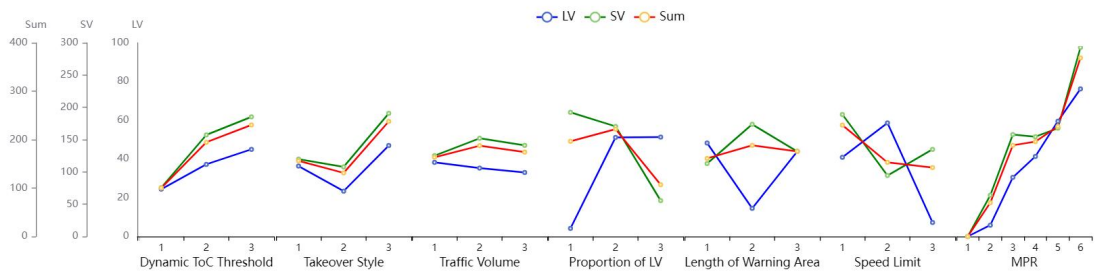

(b)Threshold

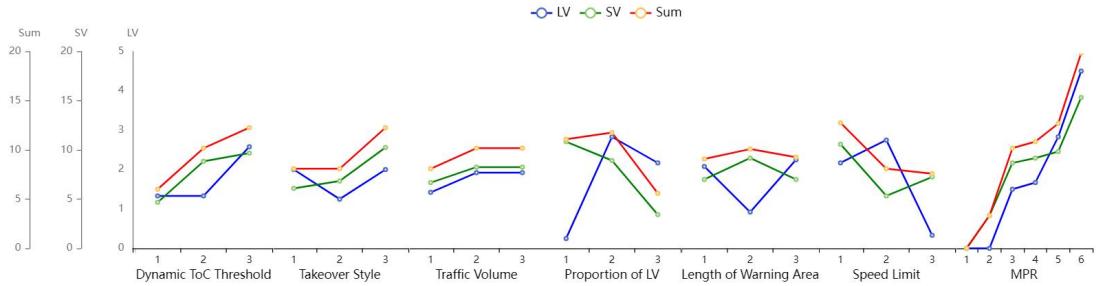

(c)MRM

Figure 5 Result of range analysis

（2）Variance Analysis

Table 10    Analysis of Variance Results

| | P (Single-vehicle conflict) | P (Multi-vehicle conflict) | P (Total conflict) | P (Disengagement of small vehicles) | P (Disengagement of large vehicles) | P (Total disengagement) | P (Small vehicles MRM) | P (Large vehicles MRM) | P (MRM total) |
|---|---|---|---|---|---|---|---|---|---|
| Intercept | 0.000** | 0.000** | 0.000** | 0.000** | 0.000** | 0.000** | 0.000** | 0.000** | 0.000** |
| Disengagement thresholds | 0.000** | 0.002** | 0.004** | 0.000** | 0.299 | 0.000** | 0.003** | 0.251 | 0.006** |
| Parameter settings for takeover styles | 0.015* | 0.000** | 0.000** | 0.000** | 0.163 | 0.002** | 0.007** | 0.787 | 0.038* |
| Traffic volume | 0.000** | 0.000** | 0.000** | 0.033* | 0.671 | 0.211 | 0.185 | 0.948 | 0.282 |
| Proportion of large vehicles | 0.038* | 0.000** | 0.000** | 0.000** | 0.001** | 0.000** | 0.000** | 0.039* | 0.002** |
| Length of warning area | 0.290 | 0.095 | 0.079 | 0.205 | 0.111 | 0.560 | 0.952 | 0.379 | 0.854 |
| Speed limit in work zone | 0.164 | 0.008** | 0.008** | 0.000** | 0.000** | 0.032* | 0.007** | 0.008** | 0.042* |
| MPR | 0.000** | 0.000** | 0.003** | 0.000** | 0.000** | 0.000** | 0.000** | 0.006** | 0.000** |

This study focuses on the following aspects through these two analyses:

● Whether automated vehicles will enhance or deteriorate safety in work zones.

- If there is an enhancement in safety due to automated driving, under what conditions can optimal safety be achieved?
- If there is a deterioration, under what circumstances will the most adverse conditions arise?
- Recommendations for traffic management.

By synthesizing the results of range analysis and variance analysis, we can draw the following conclusions:

(1) The impact of automated driving on safety

The disengagement threshold, takeover style, traffic volume, proportion of large vehicles, and MPR have a clear impact on single-vehicle conflicts, with the degree of influence ranked from highest to lowest as follows: MPR > traffic volume > disengagement threshold > takeover style > proportion of large vehicles. Similarly, the disengagement threshold, takeover style, traffic volume, proportion of large vehicles, work zone speed limit, and MPR significantly impact multi-vehicle conflicts and total conflicts, with the influence ranked as: traffic volume > takeover style > disengagement threshold > proportion of large vehicles > work zone speed limit > MPR.

When the disengagement threshold is set to Level 2, namely 10 seconds, the total number of conflicts is minimized; both excessively low and high thresholds are detrimental to safety.

When the takeover style is aggressive, the total number of conflicts is also minimized, suggesting that early takeover by the driver can reduce the occurrence of traffic conflicts, which is a reasonable conclusion.

As traffic volume increases, multi-vehicle conflicts become the predominant type, leading to a significant rise in both the number of multi-vehicle conflicts and the total number of conflicts.

The increase in the proportion of large vehicles appears to lead to a reduction in the total number of conflicts. This finding may seem counter-intuitive, as previous studies indicated that a higher number of large vehicles would decrease safety in work zones. However, our data suggest that most disengagements in work zones at similar levels of automated driving technology are instigated by small vehicles. Large vehicles, due to smaller speeds and lower acceleration and deceleration, may experience fewer disengagements. Consequently, when the proportion of small vehicles decreases, the number of traffic conflicts resulting from disengagements can also decrease to some extent.

Increasing work zone speed limits contributes to a reduction in both single-vehicle and multi-vehicle conflicts. This can be attributed to more synchronized vehicle speeds, which leads to fewer disengagements.

As MPR increases, there is a linear decreasing trend in single-vehicle conflicts, indicating that a higher proportion of automated vehicles helps reduce the occurrence of single-vehicle conflicts. Conversely, multi-vehicle conflicts and the total number of conflicts show an increasing trend with rising MPR. This illustrates that the impact of MPR on traffic conflicts is complex, diverging from previous research on fully automated driving (Abdulsattar et al. suggest that increasing MPR reduces the number of traffic conflicts[2], particularly as MPR approaches 100%, conflict numbers approach zero under such conditions). This discrepancy may be due to the presence of Level 2/3 automated driving, which leads to disengagements that disrupt traffic flow and contribute to more complex effects. As illustrated in disengagement indicator Figure 5(b), the number of disengagements rises sharply with increasing MPR, significantly affecting traffic safety.

From the above analysis, it is evident that we cannot conclude that current Level 2/3 automated driving technology contributes to enhanced safety in work zones. Under unfavorable conditions, it may even adversely impact safety in work zones.

(2) Control strategies

The disengagement threshold, takeover style, traffic volume, proportion of large vehicles, work zone speed limit, and MPR have a significant impact on small vehicle disengagements. The degree of influence is ranked from highest to lowest as follows: proportion of large vehicles > MPR > disengagement threshold > work zone speed limit > takeover style > traffic volume. For large vehicle disengagements, the proportion of large vehicles, work zone speed limit, and MPR exert a significant influence, with the impact ranked from highest to lowest as: work zone speed limit > proportion of large vehicles > MPR. Additionally, the disengagement threshold, takeover style, proportion of large vehicles, work zone speed limit, and MPR all significantly affect the total number of disengagements, with the influence ranked from highest to lowest as: MPR > disengagement threshold > proportion of large vehicles > takeover style > work zone speed limit.

The disengagement threshold, takeover style, proportion of large vehicles, work zone speed limit, and MPR have an impact on small vehicle MRM occurrences, with the degree of influence ranked from highest to lowest as follows: proportion of large vehicles > MPR > work zone speed limit > disengagement threshold > takeover style. For large vehicle MRM occurrences, the influencing factors are ranked from highest to lowest as: proportion of large vehicles > work zone speed limit > MPR. Furthermore, the disengagement threshold, takeover style, proportion of large vehicles, work zone speed limit, and MPR significantly affect the total MRM occurrences, with the influence ranked from highest to lowest as: MPR > disengagement threshold > proportion of large vehicles > work zone speed limit > takeover style.

Controlling traffic volume, takeover style, disengagement threshold, proportion of large vehicles, work zone speed limit, and MPR is essential for ensuring safety in work zones. Optimal safety can be achieved when the disengagement threshold is set to Level 2, the takeover style to Level 1, traffic volume to Level 1, the proportion of large vehicles to Level 3, and MPR to Level 1. The priority control parameters identified for enhancing work zone safety are traffic volume, disengagement threshold, and proportion of large vehicles.

## 4.4 Discussion

This study employs a comprehensive TCT for safety assessment, which addresses the limitations of traditional traffic safety assessments in work zones. As shown in the experimental results in Figure 5(a), while two and multi-vehicle conflicts are the predominant types of conflicts in work zones, single-vehicle conflicts should not be underestimated. Notably, when MPR is set to Level 1, the number of single-vehicle conflicts even surpasses that of two and multi-vehicle conflicts. This phenomenon is related to the unique characteristics of work zones, where traffic management significantly increases the risk of vehicle conflicts with fixed objects compared to ordinary road segments, aligning with previous research findings[6-9].

This research draws extensively on the work of Lücken, L., Mintsis, E., and others, utilizing their constructed ToC model for simulation studies conducted within the SUMO framework. To the authors' knowledge, this may be the first study specifically examining the effects of Level 2/3 automated driving in work zones, and the conclusions drawn may provide valuable insights for work zone safety management and traffic control. However, it is important to acknowledge the limitations of this simulation method, particularly concerning the simplistic modeling of automated disengagement. The current model only considers the distance of the vehicle from the front obstacle and the driver's reaction time, which may lead to significant discrepancies from real-world scenarios. The environment of work zones is complex, and numerous factors can trigger disengagement of automated driving, including lighting conditions, vehicle recognition of barrier facilities, road surface conditions, and weather, among others. Future studies could consider developing more complex and realistic disengagement models[40, 41] or further refining the approach using high-fidelity automated driving simulations[42].

Finally, returning to the original intent of this paper, a practical aim of the research on Level 2/3 automated driving concerning work zone safety is to propose novel management strategies. Given the current limitations of automated driving technology, which is primarily at Level 2/3 with a low MPR, the experimental results presented in this study indicate that the drawbacks outweigh the benefits regarding work zone safety. Therefore, it is essential to implement early warning measures specifically for automated vehicles within work zone management protocols. For instance, alerts should be provided to remind drivers to transition back to manual control and disengage from automated driving mode. To minimize disruptions to traffic flow caused by transitioning control, drivers should be guided to take over proactively, particularly since drivers may become distracted when their vehicles are operating in automated mode on normal road segments. This distraction can prolong both active and passive takeover times, leading to negative responses when control is required. In the near future, with advancements in technology and improvements in MPR, it should be possible to monitor the levels of automated driving technology, MPR, and driver behavior within traffic flow at work zones. Targeted control measures can then be proposed to effectively leverage automated driving capabilities to enhance and even improve safety in work zones.

# 5 Conclusion and Outlook

This study utilized traffic simulation methods to examine the impact of Level 2/3 automated driving on traffic safety in work zones, leading to the following conclusions:

1.Complex impact of automated driving technology on work zone safety: The influence of Level 2/3 automated vehicles in highway work zone environments is characterized by uncertainty. Under certain conditions, automated driving technology may enhance the safety of work zones, while in other situations, it could introduce new safety risks.

2.Significant impact of disengagement phenomena on traffic flow: The occurrence of disengagement among automated vehicles in work zones reduces the proportion of vehicles capable of maintaining automated driving status. If human takeovers are not timely or adequate, this can easily lead to new traffic conflicts, thereby becoming a new risk factor.

3.Variable influence of factors on work zone safety: Various factors, including disengagement thresholds, takeover style, traffic volume, proportion of large vehicles, work zone speed limits, and MPR, have differing degrees of impact on both single-vehicle and multi-vehicle conflicts. Notably, an increase in MPR helps to reduce the occurrence of single-vehicle conflicts, but it may also increase the likelihood of multi-vehicle conflicts.

In terms of control strategies, reducing disengagement is a critical measure, with adjustments to the proportion of large vehicles, disengagement thresholds, and MPR identified as priority control elements. Currently, automated driving technology has not demonstrated the capability to enhance traffic safety in work zones; therefore, implementing early warning systems to encourage drivers to proactively take over and disengage from automated driving mode is recommended.

In summary, while automated driving technology holds the potential to significantly improve safety levels in work zones, its practical application still faces numerous challenges and uncertainties. Consequently, future research and improvement efforts should focus on optimizing the disengagement detection and takeover mechanisms within automated driving systems. Additionally, new management approaches tailored to traffic flows that include automated vehicles should be proposed to ensure and even enhance safety levels in work zones.

# Acknowledgment

This work was sponsored by the Science and Technology Commission of Shanghai Municipality under Grant No.23YF1446000.